\documentclass[]{TEAI}
\usepackage{helvet}

\usepackage{amsmath} 
\usepackage{natbib}
\usepackage{graphicx}
\usepackage{subcaption} 
\usepackage{algorithm}
\usepackage{algorithmic}

\usepackage[toc,page,header]{appendix}
\usepackage[utf8]{inputenc} % allow utf-8 input
\usepackage[T1]{fontenc}    % use 8-bit T1 fonts
\usepackage{hyperref}       % hyperlinks
\usepackage{url}            % simple URL typesetting
\usepackage{booktabs}       % professional-quality tables
\usepackage{lmodern}        % scalable Latin Modern fonts
\usepackage{amsfonts}       % blackboard math symbols
\usepackage{nicefrac}       % compact symbols for 1/2, etc.
\usepackage{microtype}      % microtypography
\usepackage{wrapfig}

\usepackage{amssymb}  % 用于特殊符号如♠♣等
\usepackage{fontawesome}  % 用于邮箱图标 \faEnvelope
\usepackage{url}  % 用于 \url 命令

\usepackage{titletoc}

\usepackage{tikz}  % 用于绘制彩色标记符号
\usepackage{comment}  % 用于注释备用版本
\usepackage{tabularx}  % 如果需要自动调整列宽
\usepackage{booktabs}  % 用于更美观的表格线条(可选)
\usepackage{minitoc}

\usepackage{booktabs}
\usepackage{array}
\usepackage{etoolbox}

\definecolor{lightblue}{RGB}{200, 230, 255}  
\definecolor{headerblue}{RGB}{150, 200, 255} 

\usepackage{pgfplots}
\usepackage[utf8]{inputenc} % allow utf-8 input
\usepackage[T1]{fontenc}    % use 8-bit T1 fonts
\usepackage{hyperref}       % hyperlinks
\usepackage{url}            % simple URL typesetting
\usepackage{booktabs}       % professional-quality tables
\usepackage{amsfonts}       % blackboard math symbols
\usepackage{nicefrac}       % compact symbols for 1/2, etc.
\usepackage{microtype}      % microtypography
\usepackage{xcolor}         % colors
\usepackage{graphicx}
\usepackage{float}
\usepackage{comment}
\usepackage{multirow} % For multi-row cells
\usepackage{amsmath} % For \text command if needed inside math mode\Delta
\usepackage{makecell} % For multi-line cells and better vertical spacing in cells
\usepackage{siunitx}  % For better number alignment (optional but recommended)
\usepackage{tikz}
\usepackage{pgf-pie} % Package for creating pie charts
\usepackage{subcaption}
\usepackage{wrapfig}
\usepackage[export]{adjustbox}

\usepackage{ragged2e}      % for \RaggedRight in tabularx
\usepackage{tabularx}       % For tables with fixed total width and auto-adjusting columns
\usepackage{array}          % For advanced column formatting (like >{\centering\arraybackslash}X)
\usepackage{caption}        % Recommended for figures/tables, but we'll do simple text below images here.
\usepackage{enumitem}
\usepackage{pifont}
\usepackage[hang,flushmargin]{footmisc} % 更好的脚注处理

\usepackage{tcolorbox}
\usepackage{tcolorbox}    % 1. 先加载 tcolorbox 主包
\tcbuselibrary{breakable}  % 2. 显式加载 breakable 库
\tcbuselibrary{skins}      % 3. 显式加载 skins 库 (这个库提供 topruleatbreak 等选项)
\usepackage{tabularx}
\usepackage{listings}

\title{\textsc{Mirror}: A Multi-Agent System for AI-Assisted \\ Ethics Review}

\author{
Yifan Ding\textsuperscript{1},
Yuhui Shi\textsuperscript{4},
Zhiyan Li\textsuperscript{2,3},
Zilong Wang\textsuperscript{1},
Yifeng Gao\textsuperscript{1},
Yajun Yang\textsuperscript{4},
Mengjie Yang\textsuperscript{5},
Yixiu Liang\textsuperscript{6},
Xipeng Qiu\textsuperscript{1},
Xuanjing Huang\textsuperscript{1},\\
Xingjun Ma\textsuperscript{1,$\dagger$},
Yu-Gang Jiang\textsuperscript{1},
Guoyu Wang\textsuperscript{2,3,7,$\dagger$}
}

\affiliation[]{%
\parbox[t]{0.94\linewidth}{\centering
$^{1}$Institute of Trustworthy Embodied AI, Fudan University, Shanghai 200433, China;
$^{2}$Institute of Technology Ethics for Human Future, Fudan University, Shanghai 200433, China;
$^{3}$School of Philosophy, Fudan University, Shanghai 200433, China;
$^{4}$School of Life Sciences, Fudan University, Shanghai 200433, China;
$^{5}$Ethics Committee of Zhongshan Hospital, Fudan University, Shanghai 200032, China;
$^{6}$Department of Cardiology, Zhongshan Hospital of Fudan University, Institute of Cardiovascular Diseases, National Clinical Research Centre for Interventional Medicine, Shanghai 200032, China;
$^{7}$Shanghai Artificial Intelligence Laboratory, Shanghai 200030, China%
}}

\contribution[\dagger]{Corresponding authors}

\abstract{
Ethics review is a foundational mechanism of modern research governance, yet contemporary systems face increasing strain as ethical risks arise as structural consequences of large-scale, interdisciplinary scientific practice. The demand for consistent and defensible decisions under heterogeneous risk profiles exposes limitations in institutional review capacity rather than in the legitimacy of ethics oversight.
Recent advances in large language models (LLMs) offer new opportunities to support ethics review, but their direct application remains limited by insufficient ethical reasoning capability, weak integration with regulatory structures, and strict privacy constraints on authentic review materials.
In this work, we introduce \textbf{Mirror}, an agentic framework for AI-assisted ethical review that integrates ethical reasoning, structured rule interpretation, and multi-agent deliberation within a unified architecture. At its core is \textbf{EthicsLLM}, a foundational model fine-tuned on \textbf{EthicsQA}, a specialized dataset of 41,000 question–chain-of-thought–answer triples distilled from authoritative ethics and regulatory corpora. EthicsLLM provides detailed normative and regulatory understanding, enabling Mirror to operate in two complementary modes. \textbf{Mirror-ER (expedited Review)} automates expedited review through an executable rule base that supports efficient and transparent compliance checks for minimal-risk studies. \textbf{Mirror-CR (Committee Review)} simulates full-board deliberation through coordinated interactions among expert agents, an ethics secretary agent, and a principal investigator agent, producing structured, committee-level assessments across ten ethical dimensions. Empirical evaluations demonstrate that Mirror significantly improves the quality, consistency, and professionalism of ethics assessments compared with strong generalist LLMs. The framework is modular, privacy-preserving, and suitable for deployment in real institutional environments, offering a promising direction for scalable, trustworthy AI support in research ethics oversight.
}

\correspondence{\email{xingjunma@fudan.edu.cn}, \email{wguoyu@fudan.edu.cn}}
\begin{document}
\maketitle

% Catalogue (Need \newpage)
% \newpage
% \tableofcontents
% \newpage

% \vspace{-1.5em}

\section{Introduction}

In the ongoing wave of scientific and technological transformation exemplified by AI for Science, artificial intelligence has become deeply embedded across the entire research lifecycle, encompassing large-scale data acquisition anarxivd preprocessing, automated hypothesis generation, and the design of virtual experiments, with an increasing number of critical research activities delegated to algorithms and foundation models. These developments are rapidly reshaping the efficiency and epistemic scope of scientific inquiry, while substantially increasing the complexity of research objects, methodological tools, and experimental environments. Yet scientific innovation of lasting significance cannot be defined solely by technical acceleration; it must remain grounded in respect for life and the protection of human dignity. Within this context, ethical review emerges as a central pillar of science and technology governance in the AI for Science era. Beyond its function of ensuring legal and ethical compliance, ethical review serves as an institutional safeguard that conditions AI-enabled scientific research on the preservation of human dignity and the public interest, thereby providing scientific progress with both a normative foundation and a distinctly human orientation.

Ethical review is an expert-driven activity that depends critically on professional judgment and contextual reasoning. Reviewers are required, often under strict time constraints, to process large volumes of unstructured textual materials, accurately interpret the scope and applicability of relevant laws, regulations, and ethical guidelines, and formulate well-justified decisions that are defensible within specific research contexts.\cite{candilis2006need} In the AI for Science era, however, the growing scale, heterogeneity, and technical complexity of research projects are placing increasing pressure on this traditionally human-centered workflow, which relies heavily on individual experience and offline deliberation. Tasks such as manually cross-referencing multi-layered regulatory requirements, maintaining consistency of judgment under sustained cognitive load, and repeatedly drafting, revising, and communicating review comments can significantly prolong review cycles, amplify variability across decisions, and contribute to reviewer fatigue. Collectively, these limitations challenge the capacity of existing ethical review mechanisms to meet the rising demand for efficient, consistent, and sustainable ethical governance in AI-enabled scientific research.\cite{burman2001breaking, goldman1982inconsistency, mansbach2007variation, green2006impact}

In recent years, rapid advances in large language models have substantially enhanced capabilities in natural language understanding, code generation, and scientific communication, leading to their increasing deployment across a range of high-stakes domains, including clinical decision support, legal analysis, and large-scale information retrieval. \cite{cui2023chatlaw,yue2023disclawllm,liu2021caseelements,zhang2021judgmentelements,janatian2023text2structure}Owing to their ability to process extensive textual inputs and perform context-sensitive reasoning in near real time, such models demonstrate clear advantages when interacting with complex, document-intensive materials. However, despite these strengths, most existing LLMs are developed primarily for general-purpose dialogue, programming assistance, or open-domain question answering. They have not yet been systematically adapted or rigorously evaluated for the highly specialized, context-dependent, and responsibility-intensive task of ethical review, where normative interpretation, regulatory sensitivity, and accountability play a central role.

To address these gaps, we introduce \textbf{Mirror}, a multi-agent system for AI-assisted ethical review that integrates a domain-adapted foundational model, structured rule interpretation, and multi-agent deliberation within a single architecture. Mirror is designed to support both procedural compliance checks and substantive ethical reasoning while preserving the rigor and transparency expected in real-world review environments. The framework is powered by \textbf{EthicsLLM}, a specialized model for ethics reasoning, complemented by general-purpose LLMs such as Qwen3 that provide auxiliary capabilities in retrieval, classification, and coordination.

At the center of the framework is \textbf{EthicsLLM}, adapted from Qwen3-8B and fine-tuned on \textbf{EthicsQA}, a dataset of 41,000 question–chain-of-thought–answer triples distilled from more than one billion tokens of authoritative ethics literature, regulatory documents, and scholarly sources. This training equips EthicsLLM with detailed knowledge of ethical principles, regulatory structures, and expert reasoning patterns, enabling reliable rule interpretation and context-sensitive normative judgment.

Building on this foundation, Mirror operates in two modes that parallel established review procedures. \textbf{Mirror-ER (Expedited Review)} provides rapid assessment for minimal-risk studies through an executable rule base derived from statutes and institutional guidelines. Rules are expert-validated, represented as searchable graphs, and combined with retrieval-augmented analysis to generate transparent and efficient compliance reports. \textbf{Mirror-CR (Committee Review)} simulates full-board deliberation by coordinating multiple expert agents, an ethics secretary agent, and a principal investigator agent. The process involves dimension-guided evaluation across ten ethical dimensions, structured multi-agent debate, and synthesis of a committee-level assessment. Mirror-CR mirrors the dynamics of human deliberation and supports deeper, multidimensional ethical analysis.
Together, Mirror-ER and Mirror-CR enable the framework to replicate the dual-track workflow of modern ethics committees, providing a unified and adaptable system for AI-assisted ethical review.

In summary, our main contributions are as follows:

\begin{itemize}

\item We construct \textbf{EthicsQA}, a specialized dataset of 41,000 question–chain-of-thought–answer triples distilled from more than one billion tokens of authoritative ethics and regulatory sources. EthicsQA enables robust domain adaptation and equips EthicsLLM with the knowledge foundations required for expert ethical reasoning.

\item We train \textbf{EthicsLLM}, a domain-adapted foundational model fine-tuned on EthicsQA. EthicsLLM provides the conceptual, normative, and regulatory grounding necessary to support both rule-based analysis and deliberative reasoning within Mirror.

\item We propose \textbf{Mirror}, a multi-agent system for AI-assisted ethics review that supports both expedited and committee-level ethics review through two operational modes, Mirror-ER and Mirror-CR. The system combines structured rule interpretation, multi-agent deliberation, and a specialized ethics-focused foundation model.

\item We demonstrate that Mirror achieves substantial improvements in the quality, consistency, and professionalism of ethics assessments compared with general-purpose LLMs. The system remains modular and suitable for deployment in privacy-preserving institutional environments.
\end{itemize}

\section{Related Work}

\paragraph{LLM-Assisted Ethics Review.} Research on the use of LLMs to support institutional ethics review remains sparse. The most direct exploration is offered by Porsdam Mann et al.~\cite{porsdammann2025_irb_llm} who outline a conceptual roadmap for developing LLMs tailored to the needs of Institutional Review Boards through targeted fine-tuning and retrieval augmentation. Their proposal underscores the promise of LLMs in this domain but does not provide datasets, operational methodologies, or empirical demonstrations of feasibility. Complementary studies investigate whether general-purpose LLMs can identify ethical issues in medical protocols or approximate committee decisions~\cite{sridharan2024_case, sridharan2024_irb}. These results suggest initial utility but rely exclusively on off-the-shelf models, cover only a narrow set of tasks, and do not attempt to reproduce the procedural structure or deliberative depth of real ethics committees. As a consequence, the field lacks principled resources, standardized benchmarks, and end-to-end systems for AI-supported ethics review. Addressing these omissions forms a central motivation for the development of Mirror.

\paragraph{Rule-based Compliance Checking.} Automated compliance checking has been explored in various legal, regulatory, and business settings. Early systems formalize obligations and permissions using logical frameworks, workflow formalisms, or Petri nets~\cite{governatori2006_compliance, rojas2016_conformance}. These approaches provide transparent and verifiable reasoning but require substantial expert effort to encode complex regulatory language and often struggle to adapt to evolving rules or heterogeneous application scenarios. Subsequent work introduces data-driven methods that embed contractual clauses or regulatory provisions into vector representations and infer compliance through classifiers or similarity measures~\cite{aires2018_embedding, lippi2019_claudette, huang2024_multitask}. While more scalable, such methods offer limited interpretability and provide little insight into how specific provisions shape compliance outcomes. More recent hybrid frameworks integrate retrieval-augmented generation to improve flexibility and coverage~\cite{sun2025_rag_compliance}.
Although these approaches offer useful foundations, they are generally designed for relatively static regulatory environments and do not address the normative flexibility, scenario-dependent interpretation, or multistage reasoning that characterize human-subjects research ethics. Our work builds on this literature by operationalizing ethical guidelines into executable review rules while preserving transparency and adaptability.

\paragraph{Domain-Specific LLMs and Multi-Agent Systems.} Large language models increasingly benefit from domain-specific adaptation and structured multi-agent coordination. In specialized fields such as medicine and law, targeted corpus construction and instruction tuning have produced models that substantially outperform general-purpose LLMs on professional reasoning tasks~\cite{singhal2024_medpalm2, zhang2023_huatuogpt2, cui2023_chatlaw, yue2023_disc_lawllm}. Complementary work demonstrates that groups of coordinated LLM-based agents can achieve greater reliability and interpretability than isolated models. Role-conditioned collaboration, iterative critique, and workflow-guided agent organization have been shown to improve factual accuracy, reduce error propagation, and produce more coherent multi-step solutions~\cite{li2023camel, du2024multiagent, hong2023metagpt, ma2025multiagent}.
These developments illustrate the promise of domain adaptation and structured agent interaction for complex decision-making. However, institutional ethics review requires a form of reasoning that exceeds existing applications: it integrates normative analysis, evolving regulatory standards, contextual evaluation of study protocols, and committee-style deliberation across multiple disciplinary roles. No current domain-specific LLM or multi-agent framework addresses this combination of requirements. These gaps motivate the development of EthicsLLM and the design of Mirror-CR, which adapt domain specialization and multi-agent coordination to the distinctive demands of ethics review.

\section{Ethics Foundation Model}

\begin{figure*}[!t]
\centering
\includegraphics[width=\textwidth]{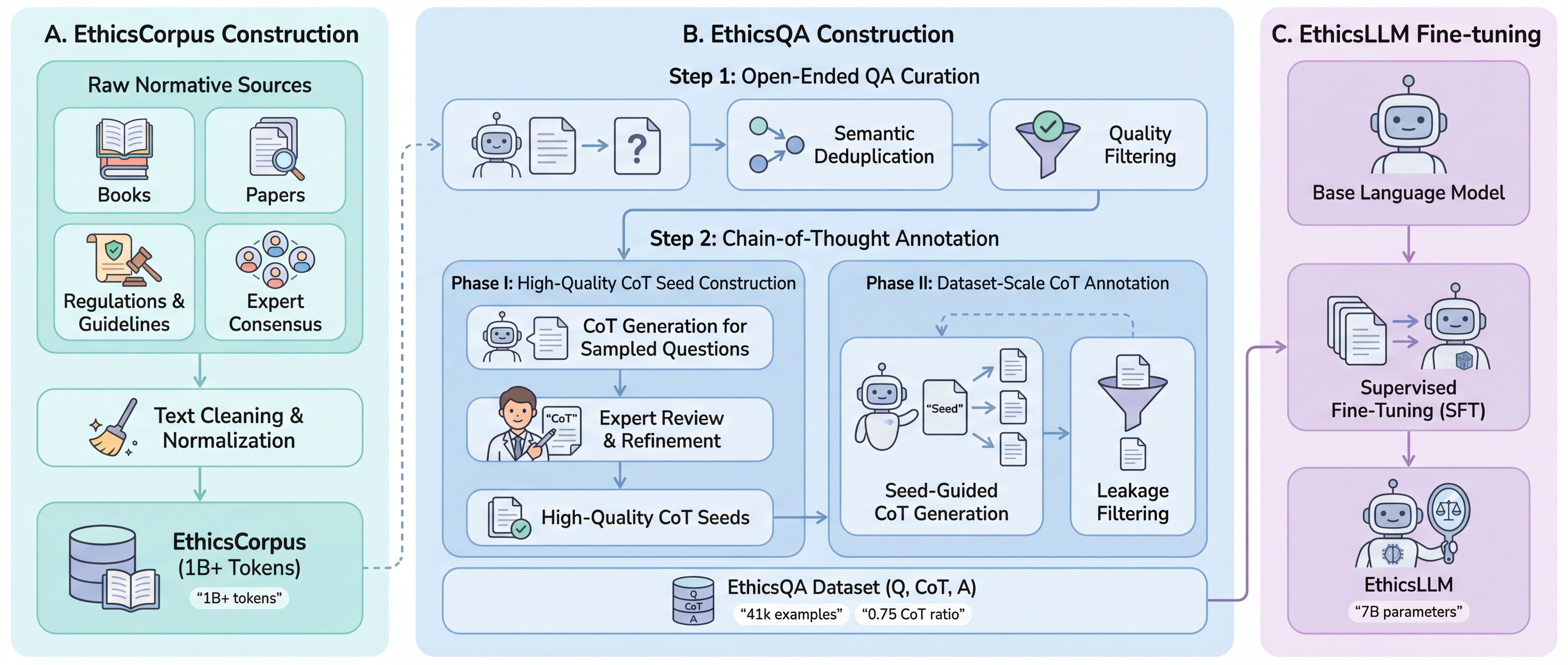}
\caption{Dataset construction and model fine-tuning pipeline for \textbf{EthicsLLM}.
(\textbf{A}) EthicsCorpus construction from diverse normative sources, followed by text cleaning and normalization.
(\textbf{B}) EthicsQA construction via open-ended QA curation and a two-phase CoT annotation process with expert-reviewed seeds and leakage filtering.
(\textbf{C}) Supervised fine-tuning of a pre-trained language model on EthicsQA to obtain an ethics-adapted foundation model for downstream ethics review.}
\label{fig:ethicsllm_pipeline}
\end{figure*}

LLMs play a central role in multi-agent systems by providing the underlying reasoning capabilities that guide coordination, interpretation, and decision-making. In the context of ethics review, this role becomes especially critical. Evaluating research involving human participants requires familiarity with normative principles, regulatory frameworks, procedural safeguards, and risk-classification criteria that extend well beyond the general linguistic and encyclopedic knowledge contained in standard foundation models. General-purpose LLMs therefore remain insufficient: they often overlook subtle ethical distinctions, misinterpret regulatory provisions, and fail to apply contextually appropriate standards of ethical judgment.

To support reliable and principled automation of ethics review, the Mirror framework incorporates \textbf{EthicsLLM}, an ethics-adapted foundation model designed to provide the normative grounding and regulatory understanding required for both rule-driven and deliberative evaluation. EthicsLLM serves as the cognitive backbone of the system, enabling consistent interpretation of ethical norms, supporting structured compliance checking, and guiding multi-agent deliberation across diverse research scenarios.

As illustrated in Fig.~\ref{fig:ethicsllm_pipeline}, EthicsLLM is obtained by fine-tuning a pre-trained LLM on \textbf{EthicsQA}, which is a supervised dataset composed of open-ended questions, CoT rationales, and answers derived from authoritative ethics and regulatory sources. EthicsQA is constructed from \textbf{EthicsCorpus}, a curated collection of normative texts that capture the conceptual, legal, and procedural foundations of contemporary human-subjects research oversight.
Next, we describe the construction process of EthicsQA and the fine-tuning procedure of EthicsLLM.
% These components establish the ethical reasoning substrate on which the entire Mirror framework operates.

\subsection{Dataset Construction}\label{sec:Dataset Construction}

To adapt a general-purpose LLM to the requirements of ethics review, we construct \textbf{EthicsQA}, a supervised dataset of $(question,\,\mathrm{CoT},\,answer)$ triples grounded in a large corpus of normative and regulatory texts. EthicsQA is designed to encode both factual ethical principles and the structured reasoning patterns that characterize human-subjects research oversight. The creation of EthicsQA proceeds through three main steps: 1) raw corpus construction, 2) open-ended question construction, and 3) chain-of-thought construction.

\paragraph{Raw Corpus Construction.}
We begin by assembling \textbf{EthicsCorpus}, a comprehensive collection of open-access documents spanning ethics books and monographs, peer-reviewed research articles, regulatory and institutional guidelines, and expert consensus statements. Documents are retrieved using expert-crafted queries and converted to plain text with the MinerU toolkit. A multi-stage cleaning pipeline removes tables, reference lists, boilerplate formatting, and OCR artifacts, yielding a corpus of over one billion tokens of high-quality normative text. Statistics of the raw corpus are provided in Table~\ref{tab:raw_corpus}. EthicsCorpus serves as the textual foundation from which EthicsQA is derived.

% \begin{table}[!t]
% \footnotesize
% \caption{Statistics of the EthicsCorpus used to construct EthicsQA.}
% \label{tab:raw_corpus}
% \tabcolsep 32pt
% \begin{tabular*}{\textwidth}{lcc}
% \toprule
% \textbf{Source} & \textbf{Number of Documents} & \textbf{Number of Tokens (M)} \\
% \midrule
% Books                     & 7{,}174  & 1{,}429 \\
% Papers                    & 14{,}630 & 133 \\
% Regulations \& Guidelines  & 538      & 14.5 \\
% Expert Consensus          & 82       & 0.53 \\
% \midrule
% \textbf{Total}    & 22{,}424 & 1{,}577 \\
% \bottomrule
% \end{tabular*}
% \end{table}
\begin{table}[!t]
\footnotesize
\caption{Statistics of the EthicsCorpus and resulting EthicsQA distribution.}
\label{tab:raw_corpus}
\tabcolsep 22pt
\begin{tabular*}{\textwidth}{lccc}
\toprule
\textbf{Source} 
& \textbf{Documents} 
& \textbf{Tokens (M)} 
& \textbf{EthicsQA Pairs} \\
\midrule
Books                     
& 7{,}174  
& 1{,}429 
& 21{,}743 \\

Papers                    
& 14{,}630 
& 133 
& 13{,}587 \\

Regulations \& Guidelines  
& 538      
& 14.5 
& 4{,}286 \\

Expert Consensus          
& 82       
& 0.53 
& 1{,}602 \\
\midrule
\textbf{Total}    
& 22{,}424 
& 1{,}577 
& 41{,}218 \\
\bottomrule
\end{tabular*}
\end{table}

\paragraph{Open-Ended Question Generation and Filtering.}
From EthicsCorpus, we generate training supervision by constructing open-ended question--answer pairs. Doubao-1.6-seed is prompted to propose several candidate questions per paragraph that probe ethical principles, regulatory obligations, or procedural requirements. Candidate questions are embedded and grouped by hierarchical clustering (cosine threshold 0.75) to eliminate redundancy. One representative question from each cluster is retained for semantic diversity. A second doubao-1.6-seed filtering stage removes questions that are ambiguous, insufficiently grounded, or reliant on external context, yielding 41{,}218 high-quality question--answer pairs. The resulting EthicsQA dataset integrates both ethical theory and institutional review practice.
It includes questions related to normative principles as well as procedural and regulatory aspects of ethics review, supporting supervision across conceptual reasoning and applied review tasks.

\paragraph{Chain-of-Thought Annotation.}
To capture reasoning patterns, we construct CoT explanations in two stages using Qwen3-32B and doubao-1.6-seed. First, Qwen3-32B is prompted in ``thinking mode'' to produce reasoning traces for a sampled subset of questions without being provided the answers. Human annotators review the generated traces and retain only those that are correct, well-structured, and aligned with ethical principles. These exemplars serve as few-shot demonstrations for CoT generation.
Next, doubao-1.6-seed is prompted to generate CoT rationales for all items, conditioned on the question and its answer, while explicitly instructed not to reveal or depend on the known answer in an obvious manner. An automated filter removes any CoT containing answer leakage, and problematic items are regenerated until no leakage occurs. The resulting \textbf{EthicsQA} dataset contains 41{,}218 high-quality $(question,\,\mathrm{CoT},\,answer)$ triples and serves as the primary supervision dataset for EthicsLLM.

\subsection{Model Fine-tuning}\label{sec:ModelTraining}

To obtain \textbf{EthicsLLM}, we fine-tune the Qwen3-8B model on EthicsQA using supervised fine-tuning (SFT). The training objective is defined as the negative log-likelihood:
\begin{equation}
\mathcal{L}_{\mathrm{SFT}} = -\frac{1}{N}\sum_{i=1}^{N}\sum_{t=1}^{|y_i|}\log P_{\theta}(y_{i,t} \mid x_i, y_{i,<t}),
\end{equation}
where $x_i$ is the question and $y_i$ is either the CoT-augmented explanation or the direct answer depending on instance type.
This training procedure equips the model with detailed ethical knowledge, regulatory interpretation skills, and the structured reasoning patterns characteristic of institutional review. The resulting \textbf{EthicsLLM} functions as a domain-adapted ethical reasoning backbone for both expedited rule-based review and multi-agent committee deliberation.

For downstream review tasks, we adopt a template-based prompting approach rather than additional task-specific fine-tuning. Template-based prompting provides a flexible mechanism for emphasizing particular ethical dimensions, highlighting relevant regulatory constraints, and adjusting the level of scrutiny according to study characteristics, such as minimal-risk surveys or interventional clinical trials. This strategy preserves the generality of EthicsLLM while enabling scenario-specific customization, making the model suitable for deployment in privacy-sensitive institutional environments where retraining is impractical.

\section{Multi-Agent Ethics Review}

% \section{Multi-Agent Ethics Review System: Mirror}

Modern ethics review operates along a dual-track process: \textit{expedited review} for studies that pose minimal risk and full \textit{committee review} for proposals requiring deeper normative analysis and multi-perspective scrutiny. An effective AI-assisted system must therefore support both pathways. Mirror is designed to align with this institutional reality by combining a rule-driven agent for expedited evaluation with a multi-agent deliberation framework that emulates committee reasoning. These two modes address fundamentally different challenges. Expedited review demands precise and scalable interpretation of regulatory rules, whereas committee review requires coordinated reasoning across disciplinary perspectives, sensitivity to contextual nuance, and the ability to surface and adjudicate competing ethical considerations. Together, they enable Mirror to provide assistance across the full spectrum of research oversight. The following subsections describe each mode in detail.

\subsection{Expedited Review}

\begin{figure*}[!t]
\centering
\includegraphics[width=\textwidth]{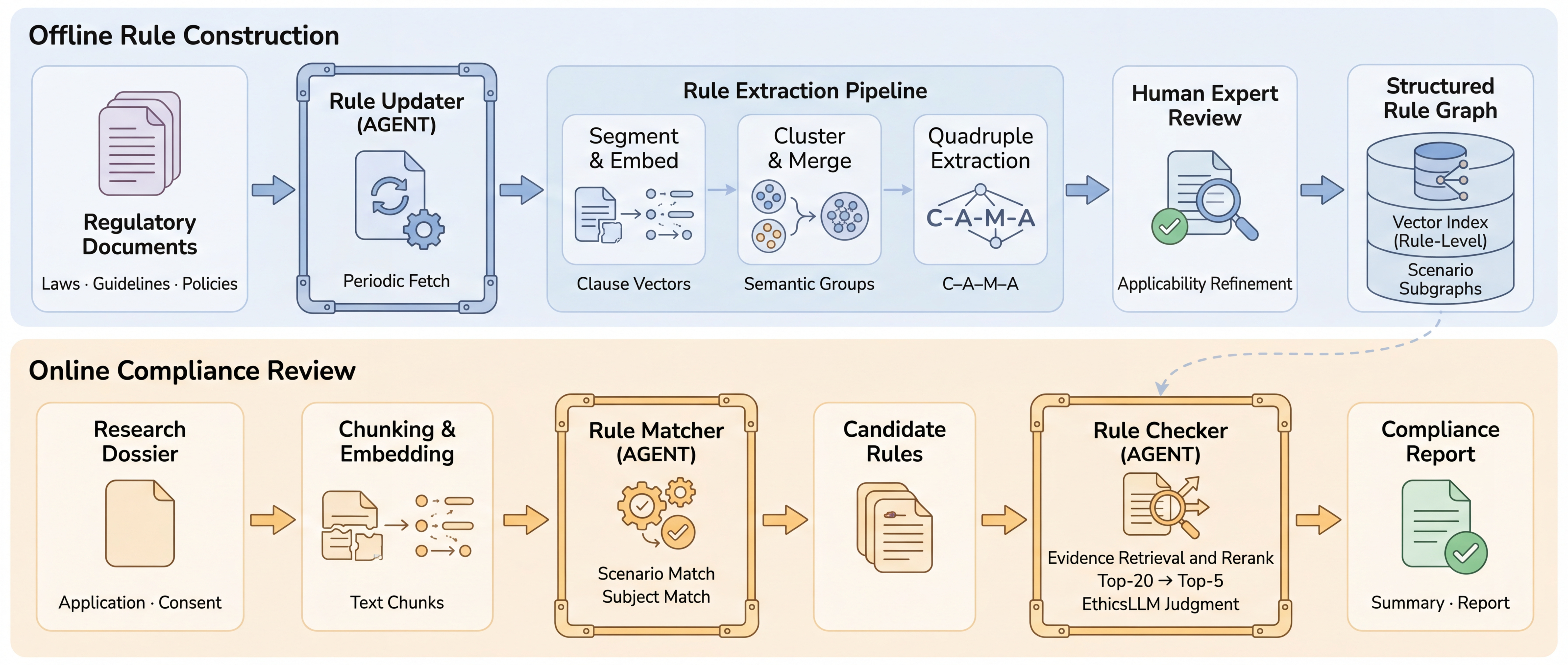}
\caption{Offline rule construction and online expedited ethics review pipeline of Mirror-ER.
In the offline stage, regulatory documents are continuously collected, canonicalized, and organized into a scenario-partitioned rule graph with expert refinement.
In the online stage, a submitted research dossier is analyzed through scenario and subject matching, followed by evidence retrieval and rule-level compliance checking with EthicsLLM, producing a structured compliance report.}
\label{fig:expedited_pipeline}
\end{figure*}

Within institutional practice, expedited review serves as a structured mechanism for evaluating minimal-risk studies through explicit regulatory and procedural criteria. Because the determinations in this pathway rely on well-defined rules rather than deliberative judgment, it provides a natural entry point for automation. We develop \textbf{Mirror for Expedited Review (Mirror-ER)} to operationalize this process through machine-executable rules, evidence-grounded analysis, and transparent compliance reporting.

The assessments of expedited review depend on determining whether an application satisfies a set of clearly articulated regulatory and procedural requirements. This rule-governed structure makes expedited review particularly amenable to computational support. 
As illustrated in Fig.~\ref{fig:expedited_pipeline}, Mirror-ER aligns submitted application materials with a continuously updated, scenario-partitioned rule graph constructed from regulatory documents, and evaluates compliance through a structured pipeline that integrates rule matching and evidence retrieval.

Built on top of \textbf{EthicsLLM}, Mirror-ER evaluates each application through a combination of rule matching, evidence retrieval, and structured reasoning. The agent produces transparent, itemized compliance reports and maintains an interpretable workflow that can be continuously updated as policies evolve. This design enables scalable support for high-volume oversight while preserving the rigor and consistency required of institutional ethics review.

\subsubsection{Rule Extraction}

Automated expedited review requires a rule base that accurately captures the structure of relevant ethical guidelines, regulatory requirements, and institutional policies. To construct such a rule base, we design a pipeline that converts heterogeneous regulatory texts into standardized, canonical rules. Each rule is expressed as a quadruple:
\[
(condition,\; subject,\; deontic\; word,\; action),
\]
which specifies its applicability criteria, responsible entity, normative status, and required behavior.

\begin{figure*}[!t]
\centering
\includegraphics[width=\textwidth]{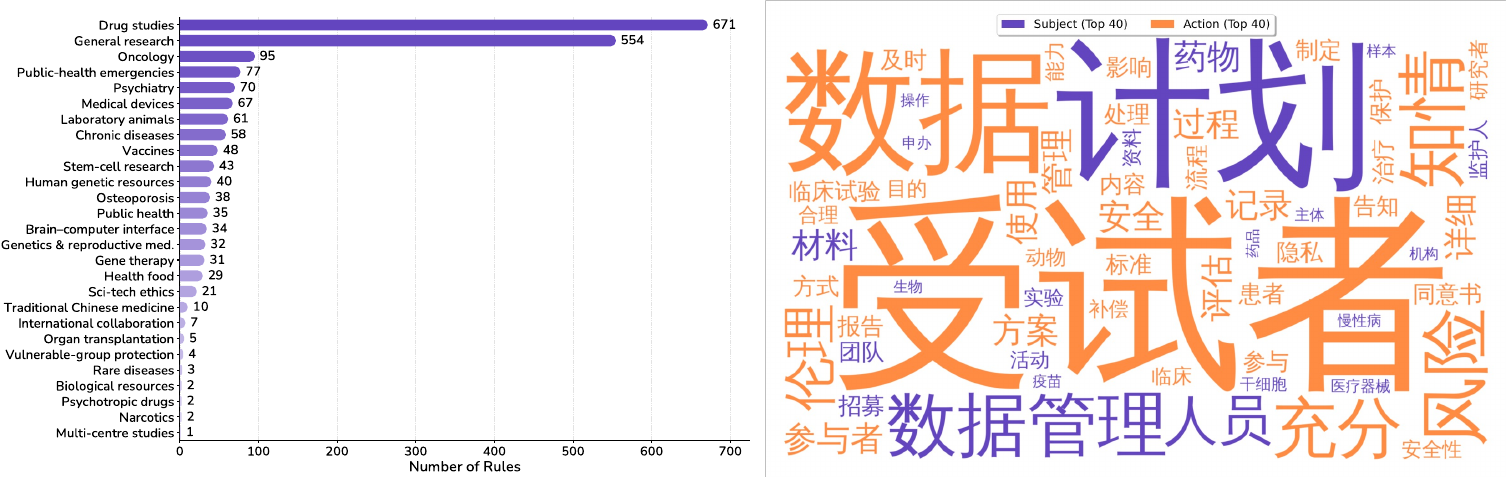}
\caption{Statistical analysis of rule conditions and semantic components in expedited ethics review.
(\textbf{Left}) Distribution of extracted rule conditions across research domains.
(\textbf{Right}) Word cloud visualization of the most frequent terms associated with rule semantics, including regulatory subjects and regulated actions. 
Within each semantic category, word sizes are normalized by their relative frequency among the top-$40$ terms in that category, highlighting the internal composition and emphasis of rule semantics rather than absolute frequency differences across categories.}
\label{fig:rule_statistics}
\end{figure*}

\paragraph{Rule Extraction.} The pipeline segments documents into clauses, extracts atomic rule candidates, embeds each candidate for semantic comparison, and clusters candidates that represent equivalent obligations. Each cluster is then merged into a single canonical rule and mapped to the quadruple form, yielding a comprehensive and machine-executable rule base (detailed in Algorithm~\ref{alg:rule_construct}). 
As shown in Fig.~\ref{fig:rule_statistics}, the resulting rule base spans a broad range of research domains and exhibits structured semantic regularities in both regulatory subjects and regulated actions, reflecting the diversity and internal composition of real-world ethics requirements.
Because regulatory landscapes evolve, Mirror-ER incorporates a rule updater that periodically retrieves newly issued policies and processes them using the same construction pipeline. This mechanism keeps the rule base aligned with current standards without requiring model retraining. Applicability conditions remain the most error-prone component, so domain experts review and refine these fields to ensure correct scope and prevent misapplication across research contexts.

\begin{algorithm}[t]
\footnotesize
\caption{Rule Extraction}
\label{alg:rule_construct}
\begin{algorithmic}[1]

\REQUIRE Regulatory documents $\mathcal{D} = \{ d_1, d_2, \dots, d_{|\mathcal{D}|} \}$
\ENSURE Executable rule set $\mathcal{E}$ in quadruple form $(cond, subject, dword, action)$

\STATE $\mathcal{E} \Leftarrow \varnothing$  \COMMENT{initialize rule base}
\STATE buffer $\Leftarrow \varnothing$      \COMMENT{global embedding buffer}

\FORALL{document $d$ in $\mathcal{D}$}
    \STATE $C \Leftarrow \textsc{SegmentIntoClauses}(d)$ 
    \FORALL{clause $c$ in $C$}
        \STATE $\mathcal{R} \Leftarrow \textsc{ExtractCandidateRules}(c)$ 
        \COMMENT{produce atomic, decoupled candidate rules}
        \FORALL{ruleCandidate $r$ in $\mathcal{R}$}
            \STATE $\mathbf{v}(r) \Leftarrow \textsc{SentenceEmbedding}(r)$ 
            \STATE Add $\mathbf{v}(r)$ to buffer
        \ENDFOR
    \ENDFOR
\ENDFOR

\STATE $\mathcal{G} \Leftarrow \textsc{HierarchicalClustering}(\text{buffer}, 0.75)$ 
\COMMENT{group semantically equivalent rules}

\FORALL{cluster $g$ in $\mathcal{G}$}
    \STATE $r^\star \Leftarrow \textsc{MergeRules}(g)$ 
    \COMMENT{semantic fusion into a canonical rule}
    \STATE $(cond, subject, dword, action) 
           \Leftarrow \textsc{ExtractQuadruple}(r^\star)$
    \STATE Add $(cond, subject, dword, action)$ to $\mathcal{E}$
\ENDFOR

\RETURN $\mathcal{E}$

\end{algorithmic}
\end{algorithm}

\paragraph{Rule Graph Construction.}
Once the canonical rule set is obtained, we organize it into a structured \textbf{rule graph} that supports systematic traversal and downstream compliance analysis. In the rule graph, nodes correspond to regulatory subjects (e.g., investigators, institutions, participants) and regulated actions, while directed edges encode their deontic relationships under specific applicability conditions.
To prevent inappropriate cross-domain inference, rules are first partitioned into \emph{scenario-specific subgraphs} that reflect distinctions among research contexts, such as biomedical experimentation, social-behavioral studies, or data-driven research. Within each subgraph, rules are encoded at the \emph{rule level} and indexed to support efficient access and interpretation.
Although rule canonicalization reduces redundancy, applicability conditions may remain ambiguous or overly broad. We therefore incorporate a \emph{human-in-the-loop} refinement step to clarify rule conditions and ensure accurate scenario partitioning. This refinement improves the structural integrity of the rule graph and supports reliable downstream review.

\subsubsection{Rule Matching}
Rule matching aims to ensure \emph{comprehensive coverage} of the rules retrieved from the rule graph. Unlike typical information-retrieval tasks that prioritize precision, ethics review places a strong emphasis on recall: any rule that may plausibly apply to a submission must be examined to avoid overlooking potential ethical risks.
Given an ethics application, the \textit{Rule Matcher} agent first identifies the relevant research scenario by analyzing project metadata and core descriptors, such as study type, participant population, and intervention characteristics. Based on this classification, the corresponding scenario-specific subgraph is selected.
Within the selected subgraph, the Rule Matcher further identifies the regulatory subjects involved in the submission (e.g., investigators, participants, institutions, data controllers). All rules associated with the identified subjects are then retained for review. Importantly, \emph{all} rules relevant to the selected scenario and subjects are forwarded to the subsequent checking stage. No similarity-based pruning or top-$k$ selection is applied. This design reflects the risk-sensitive nature of ethics review, where missing an applicable obligation is more costly than examining additional rules.

\subsubsection{Rule Checking}
To determine whether a submission satisfies or violates each applicable rule, we introduce a \textit{Rule Checker} agent that performs fine-grained, evidence-based compliance assessment.
For each rule identified by the Rule Matcher, the agent retrieves supporting evidence from the application materials using a retrieval-augmented pipeline. Specifically, the rule description is embedded and used as a query to search the project documents for relevant passages. The top twenty candidate segments are first retrieved and then re-ranked with a cross-encoder to select the five most informative evidence spans.These passages, together with the canonical rule and a domain-specific evaluation prompt, are then provided to EthicsLLM to generate an evidence-grounded judgment. The model outputs a binary compliance verdict, a natural-language rationale, and references to the supporting evidence.

Finally, Mirror-ER aggregates the above rule-level evaluations into a structured report that includes an executive summary highlighting major risks and a detailed appendix documenting assessments for every applicable rule. This workflow produces transparent, reproducible, and interpretable expedited review results.

\subsection{Committee Review}

\begin{figure*}[!t]
\centering
\includegraphics[width=\textwidth]{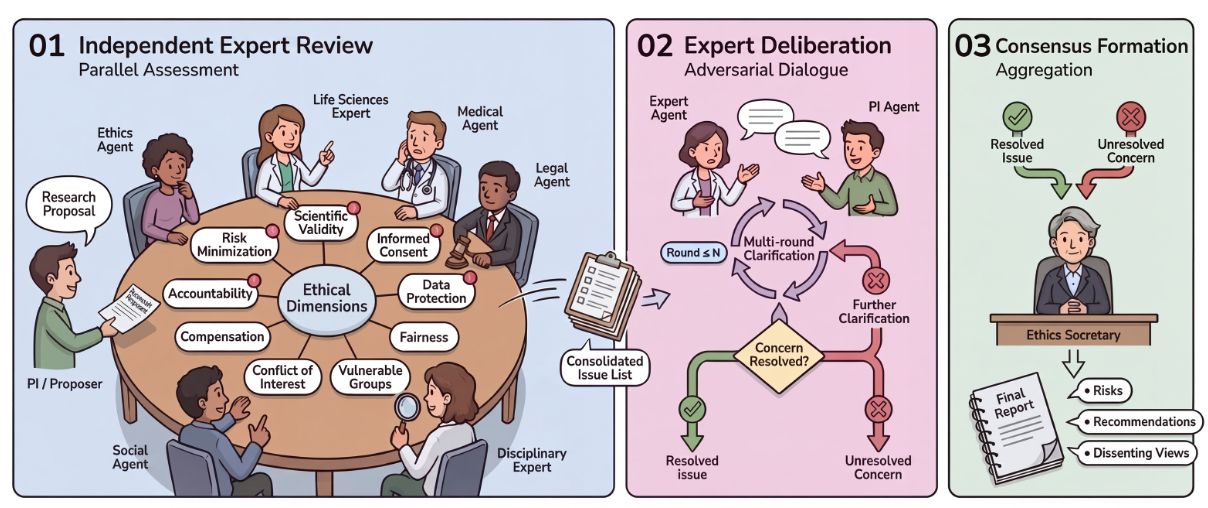}
\caption{Multi-agent committee review workflow in \textbf{Mirror-CR}.
(\textbf{01}) \textbf{Dimension-guided review}: multiple expert agents independently assess the research proposal across ten core ethical dimensions, producing a consolidated issue list.
(\textbf{02}) \textbf{Multi-agent debate}: expert agents and the PI agent engage in iterative clarification rounds to resolve raised concerns; unresolved issues persist after a bounded number of rounds.
(\textbf{03}) \textbf{Committee synthesis}: an ethics secretary agent aggregates both resolved and unresolved issues into a structured final report, including key risks and recommendations.}
\label{fig:committee_review}
\end{figure*}

While expedited review addresses studies with minimal risk, many research protocols present ethical complexities that cannot be resolved through rule-based evaluation alone. These cases require holistic, multi-perspective deliberation that characterizes full committee review. To support such scenarios, we instantiate \textbf{Mirror} in a committee-review mode (\textbf{Mirror-CR}) that emulates the workflow, division of expertise, and structured reasoning processes of institutional ethics committees. In this mode, Mirror coordinates multiple expert agents, an ethics secretary agent, and a simulated principal investigator (PI) to reproduce the collective judgment essential to committee-based evaluation.

As illustrated in Fig.~\ref{fig:committee_review}, Mirror-CR operates through three sequential stages. Expert agents first conduct dimension-guided assessments spanning key ethical domains. This is followed by a structured debate between experts and the PI, during which arguments are clarified, challenged, and refined. Finally, the ethics secretary synthesizes the deliberation into a coherent committee-level decision. This staged design captures both the procedural rigor and deliberative depth of real-world ethics review while preserving transparency and reproducibility.

\subsubsection{Ethics Committee Simulation}

Mirror-CR models the composition of an institutional ethics committee in accordance with the regulatory requirements specified in the \emph{Measures for the Ethical Review of Life Science and Medical Research Involving Humans}\cite{NHWC2023EthicsReviewEn}. The regulation stipulates that ethics review committees shall include experts from life sciences, medicine, ethics, and law, as well as experts from relevant research fields and non-institutional social representatives, with consideration given to gender diversity.

Based on these requirements, Mirror-CR instantiates a set of expert agents corresponding directly to the legally mandated categories. An \textbf{Ethics Reviewer} represents expertise in bioethics and evaluates compliance with fundamental ethical principles. A \textbf{Life Sciences Expert} assesses scientific validity, biosafety considerations, and experimental integrity. A \textbf{Medical Expert} examines issues related to participant protection, clinical risk, and potential health impacts. A \textbf{Legal Expert} interprets applicable laws, administrative regulations, and normative documents governing research involving human participants. A \textbf{Disciplinary Expert} evaluates methodological rigor and domain-specific research standards. A \textbf{Social Representative}, reflecting the role of non-institutional members required by regulation, considers issues of fairness, societal impact, and the protection of vulnerable populations.
In accordance with the regulatory emphasis on committee diversity, the gender attribute of each expert agent is assigned randomly. An \textbf{Ethics Secretary} is introduced as a procedural support role responsible for organizing the review process and recording deliberation outcomes, without participating as a voting member. The \textbf{Project Principal Investigator (PI)} participates in the review process as the subject of evaluation, providing clarifications and responses to issues raised by the committee.

\subsubsection{Multi-stage Committee Deliberation}

The committee deliberation process in Mirror-CR is organized into three sequential stages that correspond to distinct functions commonly observed in institutional ethics review procedures. These stages separate individual expert judgment, interactive examination of contested issues, and collective decision formulation, thereby enabling a structured progression from issue identification to committee-level determination.

\paragraph{Phase I: Independent Expert Review.}
The deliberation process begins with an independent evaluation conducted by each expert agent across ten predefined ethical dimensions, including risk minimization, scientific validity, informed consent, data protection, justice, vulnerability, conflicts of interest, compensation, accountability, and cross-jurisdictional coordination. During this stage, expert agents operate independently and without interaction, applying domain-specific criteria and normative guidance provided by EthicsLLM. The outcome of this stage is a structured list of ethical concerns identified by individual experts and designated for subsequent examination.

\paragraph{Phase II: Expert Deliberation.}
Ethical concerns identified during the independent review stage are subjected to structured deliberation involving the relevant expert agent and the project PI. For each issue, the expert articulates the ethical basis and potential implications of the concern, while the PI provides clarification, justification, or proposed mitigation measures. The deliberation proceeds through multiple interaction rounds, with prior arguments retained to ensure continuity of reasoning. Issues are considered provisionally resolved when the expert determines that the response adequately addresses the underlying ethical concern; unresolved issues are retained for committee-level consideration.

\paragraph{Phase III: Consensus Formation.}
Following the completion of expert deliberations, the ethics secretary consolidates all unresolved issues into a committee-level assessment. This stage focuses on synthesizing individual expert positions into an integrated evaluative outcome, explicitly attributing each concern to its corresponding domain of expertise. The resulting assessment summarizes substantive ethical risks and formulates targeted recommendations for protocol revision, additional safeguards, or further review (with new/updated supporting documents). Where consensus cannot be reached, divergent expert views are documented to preserve the plurality of ethical judgments.

\section{Experiments}
In this section, we first describe the experimental setup and then present evaluation results across different evaluation tasks, including ethics QA, rule-based expedited review, and case studies of committee review.

\subsection{Experimental Setup}

\textbf{EthicsLLM Training.}\;
We train our ethics foundation model \textbf{EthicsLLM} via full-parameter supervised fine-tuning (SFT) initialized from the \textbf{Qwen3-8B} backbone. Training is performed for three epochs using a cosine learning rate schedule with a peak learning rate of $2\times10^{-5}$ and a 10\% warm-up phase. To enable efficient large-scale optimization, we employ DeepSpeed ZeRO-3 for optimizer state and gradient sharding.
To improve robustness and adaptability in ethical reasoning, we adopt a \textbf{dual-mode training strategy} that interleaves CoT and non-CoT supervision within each batch. Non-CoT samples require direct answer prediction without intermediate reasoning, whereas CoT samples include explicit reasoning chains grounded in ethical principles and regulatory interpretation. This hybrid training formulation encourages the model to internalize both concise decision-making patterns and structured, multi-step ethical reasoning.
The global batch size is set to 256, with a maximum sequence length of 8k tokens. All experiments are conducted on eight NVIDIA A800 GPUs (80GB) using the \texttt{trl} framework. During inference, unless otherwise specified, we adopt deterministic decoding with temperature set to 0.1 to ensure stable and reproducible outputs.

\paragraph{Evaluation Tasks.}
We evaluate EthicsLLM and Mirror across three representative ethics-review tasks that reflect distinct stages of real-world institutional review workflows: (1) ethics QA, (2) expedited rule-based ethics review, and (3) committee-level ethical deliberation. Each task differs in its reasoning requirements, evaluation protocol, and appropriate comparison baselines, enabling a comprehensive assessment of both foundational ethical reasoning and system-level deliberation capabilities.

\emph{Task 1: Ethics QA}\;.This is an ethics question answering task designed to evaluate a model’s ability to interpret ethical principles, regulatory requirements, and scenario-based dilemmas. Our evaluation is performed on a unified benchmark of 8,498 questions drawn from two sources: the held-out split of \textbf{EthicsQA}, which targets general research ethics, and \textbf{EthicsQA-ER}, which focuses on regulatory interpretation in ethics review settings. Together, these datasets cover conceptual understanding, procedural compliance, and applied ethical reasoning. 
In addition to ethics-specific evaluation, we include two widely used general-capability benchmarks, \textbf{MMLU-Redux}\cite{gema2025_mmlu} and \textbf{C-Eval}\cite{huang2023ceval}, to assess whether ethics-oriented fine-tuning degrades the model’s broader reasoning and knowledge abilities. MMLU-Redux evaluates multidisciplinary understanding across a wide range of academic subjects, while C-Eval focuses on comprehensive knowledge and reasoning skills in the Chinese language context.
We evaluate EthicsLLM in both CoT and non-CoT settings to assess its performance under concise decision-making and explicit reasoning paradigms.

\emph{Task 2: Expedited Ethics Review.}\;
The expedited review task focuses on deterministic, rule-level compliance assessment for minimal-risk studies. Given its emphasis on transparency and reproducibility, this task is evaluated exclusively under non-CoT inference. We curate a benchmark of 100 fully de-identified real-world research ethics dossiers, each annotated by domain experts with violated rules, supporting evidence spans, and explanatory rationales. This setting enables fine-grained evaluation of violation detection accuracy and explanation quality under realistic institutional constraints.

\emph{Task 3: Committee Review.}\;
The committee review task targets high-risk or ambiguous research proposals that require holistic ethical judgment beyond executable rules. Due to the lack of publicly available IRB deliberation records, we construct a qualitative evaluation set of 10 synthetic but realistic research proposals authored by senior ethics experts. These cases intentionally incorporate non-rule-based dilemmas, including ambiguous risk classification, conflicting ethical obligations, cross-domain concerns, and vulnerable populations. System outputs are evaluated through expert-driven qualitative analysis rather than quantitative metrics.

\paragraph{Comparison Baselines.}

\emph{Ethics QA.}\;
For non-CoT evaluation, we compare EthicsLLM against \textbf{Qwen3-8B}, \textbf{Qwen3-14B} and \textbf{Qwen3-32B} executed in \texttt{no\_think} mode, as well as the commercial model \textbf{GPT-4.1}. For CoT evaluation, baselines include the same Qwen models in \texttt{think} mode and \textbf{DeepSeek-R1}, a state-of-the-art open-source reasoning model. All models are evaluated with internet access disabled to ensure a controlled comparison.

\emph{Expedited Ethics Review.}\;
EthicsLLM is compared against \textbf{Qwen3-8B}, \textbf{DeepSeek-V3} and \textbf{GPT-4.1}. To isolate reasoning and rule-interpretation ability, all models are provided with identical executable rule graphs, retrieved evidence passages, and structured prompts. This design controls for retrieval quality and prompt engineering effects.

\emph{Committee Review.}\;
We compare \textbf{Mirror-CR} against two strong single-agent baselines: \textbf{DeepSeek-R1} and \textbf{GPT-4.1}. Each baseline acts as an individual expert reviewer with access to the full proposal dossier and is prompted using the same instruction template as the committee system. This controlled setup isolates the effect of multi-agent deliberation from that of single-agent reasoning under identical information conditions.

\paragraph{Performance Metrics.}
Evaluation metrics are tailored to the goals and characteristics of each task.

\emph{Ethics QA.}\;
The primary metric is accuracy. All questions are answered using a fixed output template and deterministic decoding (temperature 0.1). Multiple-choice, true/false, and cloze-style questions are scored as binary outcomes. Open-ended responses are evaluated by the \texttt{doubao-1.6-seed} reviewer, and the resulting scores are linearly normalized to the $[0,1]$ range to ensure comparability across question types.

\emph{Rule-Based Ethics Review.}\;
We evaluate system performance along two complementary dimensions based on expert evaluation: (1) \emph{Quality} assesses the accuracy of rule-violation identification using \emph{precision}, \emph{recall}, and \emph{F1 score}. 
\emph{Precision} measures whether the violations flagged by the system are supported by the regulatory rules and submission evidence, reflecting the system’s ability to avoid unsupported or spurious findings. 
\emph{Recall} measures the extent to which applicable rule violations are successfully identified, which is particularly critical in ethics review where missing a relevant obligation may lead to unaddressed ethical risks. 
\emph{F1} score summarizes the trade-off between these two aspects and provides an overall indicator of violation-detection reliability. 
Consistent with institutional practice, our evaluation places stronger emphasis on recall, as comprehensive coverage is prioritized over aggressive pruning in risk-sensitive review settings; and (2) \emph{Professionalism} evaluates the quality of the generated review reports from the perspective of institutional ethics committees. This dimension captures clarity, completeness, and expert-likeness, including whether the report presents structured reasoning, appropriate regulatory references, and a professional review tone. 

\emph{Committee Review (Qualitative).}\;
Full-board ethical deliberation involves open-ended, context-sensitive reasoning and does not admit objective quantitative metrics. We therefore evaluate Mirror-CR through structured case studies. Ethics experts compare the outputs of Mirror-CR, GPT-4.1, and DeepSeek-R1 side by side and provide free-form assessments of whether each system identifies salient ethical issues, conducts coherent deliberation, and produces practically actionable recommendations. This qualitative protocol reflects standard practice in evaluating systems designed for high-level ethical judgment.

\subsection{Ethics QA Results}

\begin{table*}[t]
\centering
\footnotesize
\caption{Ethics QA and general capability accuracy (\%).
Models are evaluated under \emph{No-Thinking} (CoT disabled) and \emph{Thinking} (CoT enabled).
Best scores are shown in \textbf{bold}, and second-best scores are \underline{underlined}.}
\label{tab:qa_results}
\setlength{\tabcolsep}{4.5pt}
\begin{tabular*}{\textwidth}{@{\extracolsep{\fill}}l l cccc}
\toprule
\textbf{Model} & \textbf{Mode} &
\multicolumn{2}{c}{\textbf{Ethics QA} $\uparrow$} &
\multicolumn{2}{c}{\textbf{General Capability} $\uparrow$} \\
\cmidrule(lr){3-4} \cmidrule(lr){5-6}
 &  & \textbf{EthicsQA} & \textbf{EthicsQA-ER} & \textbf{MMLU-Redux} & \textbf{C-Eval} \\
\midrule

Qwen3-8B      & \multirow{5}{*}{No-Thinking}
              & 65.40 & 53.95 & 79.52 & 77.91 \\
Qwen3-14B     & 
              & 67.40 & 57.86 & 82.04 & \underline{81.04} \\
Qwen3-32B     & 
              & \underline{73.80} & \underline{64.73} & \underline{85.73} & \textbf{83.32} \\
GPT-4.1       & 
              & 73.20 & 62.21 & \textbf{92.43} & 77.97 \\
\textbf{Mirror (8B, Ours)} 
              & 
              & \textbf{76.07} & \textbf{66.06} & 79.13 & 77.08 \\
\midrule

Qwen3-8B      & \multirow{5}{*}{Thinking}
              & 63.60 & 56.67 & 87.52 & 83.45 \\
Qwen3-14B     & 
              & 67.27 & 59.68 & 88.61 & 86.26 \\
Qwen3-32B     & 
              & \underline{73.80} & \underline{67.69} & \underline{90.92} & \underline{87.33} \\
DeepSeek-R1 (671B)   & 
              & \textbf{83.20} & \textbf{82.12} & \textbf{92.91} & \textbf{91.82} \\
\textbf{Mirror (8B, Ours)} 
              & 
              & 69.33 & 62.94 & 86.37 & 82.17 \\
\bottomrule
\end{tabular*}
\end{table*}

Table~\ref{tab:qa_results} reports model accuracy on ethics-oriented question answering benchmarks together with two general capability benchmarks under both \emph{No-Thinking} and \emph{Thinking} inference modes. The results show that targeted ethics-domain supervised fine-tuning substantially improves domain-specific performance, enabling an 8B model to compete effectively with significantly larger LLMs, while largely preserving general reasoning capability.

\paragraph{No-Thinking Performance.}
In the direct-answer setting, \textbf{Mirror (8B)} achieves the strongest performance on both ethics benchmarks, obtaining 76.07\% on \textbf{EthicsQA} and 66.06\% on \textbf{EthicsQA-ER}. These results surpass all baselines, including Qwen3-32B and GPT-4.1, despite Mirror’s substantially smaller parameter count. This demonstrates that ethics-focused supervised fine-tuning is highly effective at enhancing factual comprehension and regulatory interpretation. On general capability benchmarks, Mirror attains 79.13\% on MMLU-Redux and 77.08\% on C-Eval, remaining comparable to Qwen3-8B and indicating that ethics specialization does not incur a noticeable degradation in general knowledge or reasoning ability.

\paragraph{Thinking Performance.}
With CoT reasoning enabled, DeepSeek-R1 (671B MoE) achieves the best overall performance, reflecting the advantage of larger models with multi-step reasoning capability. Compared with the \emph{No-Thinking} mode, \textbf{Mirror} exhibits a different performance profile under CoT mode: while ethics-domain accuracy remains higher than that of its 8B backbone, the relative gains are smaller, and general capability scores show a more pronounced decline. In particular, Mirror remains close to Qwen3-8B on MMLU-Redux and C-Eval, but the margin to larger models widens compared with direct-answer inference, and improvements on EthicsQA and EthicsQA-ER are more modest than expected.
We attribute this behavior to a distribution shift between the distilled CoT annotations used during training and the native reasoning patterns of the base Qwen model. Although the distilled CoT data provides structured ethical reasoning signals, residual stylistic and structural mismatches may limit the effectiveness of CoT inference and partially offset its potential benefits.

These results demonstrate Mirror’s strong parameter efficiency and robust domain adaptation capability. It delivers state-of-the-art performance among comparable-scale models on ethics QA tasks in the No-Thinking mode, while maintaining competitive—though more conservative—performance under CoT reasoning. These findings validate the effectiveness of the ethics-focused training pipeline and highlight Mirror’s suitability as a lightweight yet capable foundation model for downstream ethics review agents.

\begin{table*}[!t]
\centering
\footnotesize
\caption{Rule-based ethics-review results.
Quality metrics (Recall, Precision, F1) and \textbf{Professionalism} are reported in $[0,1]$ (higher $\uparrow$ is better).
The best values are shown in \textbf{bold}, and second-best values are \underline{underlined}.}
\label{tab:review_results}
\tabcolsep 6pt
\begin{tabular*}{\textwidth}{@{\extracolsep{\fill}}l l ccc c}
\toprule
\textbf{Model} & \textbf{Method} &
\multicolumn{3}{c}{\textbf{Quality} $\uparrow$} &
\textbf{Professionalism} $\uparrow$ \\
\cmidrule(lr){3-5}
 & & Recall$ \uparrow$ & Precision$ \uparrow$ & F1$ \uparrow$ & \\
\midrule

DeepSeek-V3 (671B) & \multirow{3}{*}{Zero-shot}
            & 0.1806 & 0.3333 & 0.2342 & 0.2000 \\
GPT-4.1     & 
            & 0.1528 & 0.2200 & 0.1803 & 0.4083 \\
Qwen3-8B    & 
            & 0.1231 & 0.1842 & 0.1476 & 0.1500 \\
\midrule

DeepSeek-V3 (671B) & \multirow{4}{*}{Mirror-Agent}
            & 0.4583 & \textbf{0.8462} & 0.5946 & 0.2000 \\
GPT-4.1     & 
            & \underline{0.6319} & 0.6791 & \underline{0.6547} & 0.4000 \\
Qwen3-8B    & 
            & 0.5523 & 0.6063 & 0.5780 & 0.2500 \\
\textbf{Mirror (8B, Ours)} 
            & 
            & \textbf{0.9444} & \underline{0.7640} & \textbf{0.8447} & \textbf{0.7917} \\
\bottomrule
\end{tabular*}
\end{table*}

\subsection{Expedited Review Results}
\label{sec:review_results}

Table~\ref{tab:review_results} reports the performance of different models on the rule-based ethics-review benchmark under two evaluation settings: (i) a \emph{zero-shot} setting, in which each dossier is directly assessed by a general-purpose LLM without structured guidance, and (ii) the \emph{Mirror-Agent} setting, which applies the full expedited review pipeline, including rule graph construction, rule matching, evidence retrieval, and rule-grounded evaluation.

\paragraph{Zero-shot Setting.}
In the zero-shot condition, all models exhibit limited capability in identifying regulatory violations. DeepSeek-V3 achieves the highest F1 score (0.2342), while GPT-4.1 attains the highest professionalism score, reflecting more fluent but weakly grounded explanations. Qwen3-8B performs the weakest across all metrics. Overall, these results indicate that general-purpose LLMs struggle to reliably align free-text research descriptions with specific regulatory requirements when no structured rule context is provided.

\paragraph{Mirror-Agent Setting.}
When evaluated within the full Mirror-Agent framework, all baseline models show substantial improvements in violation detection quality, confirming the importance of structured rule grounding for expedited ethics review. DeepSeek-V3 and GPT-4.1 both achieve large gains in recall and F1, demonstrating that the Mirror-Agent pipeline can significantly stabilize and guide model predictions even when paired with general-purpose LLMs. Qwen3-8B also benefits from the agentic workflow, though its overall performance remains below that of larger models. Our proposed \textbf{Mirror} agent framework with the ethics-adapted foundation model achieves the strongest overall performance. \textbf{Mirror} attains the highest recall (0.9444), F1 score (0.8447), and professionalism score (0.7917), substantially outperforming all baseline models. The high recall indicates effective coverage of potential violations, while the relatively balanced precision suggests that these gains do not come at the cost of excessive false positives. These results demonstrate that combining domain-adapted ethics knowledge with structured rule representations and an agentic review pipeline is crucial for producing accurate, consistent, and professional expedited ethics-review outcomes.

The above results indicate that (i) zero-shot ethics review remains challenging for general-purpose LLMs, (ii) structured agentic review pipelines are essential for reliable ethics review, and (iii) our proposed Mirror provides an effective and practical solution for expedited ethics review.

\subsection{ Case Studies of Committee Review}
\label{sec:committee_cases}

To further examine the capability of Mirror in high-level ethical deliberation, we conduct a committee-style evaluation on ten synthetic dossiers constructed by domain experts. These dossiers are designed to contain \emph{subtle, non-rule-based ethical risks} that are difficult to capture through explicit regulatory clauses. Since no ground-truth labels exist for such open-ended deliberation, we present two representative case studies comparing the performance of Mirror-Committee-Review with GPT-4.1 and DeepSeek-V3 under identical committee configurations.

\paragraph{Case 1: Familial Genetic Disorder Investigation.}
The first dossier describes a project conducting a familial survey of a hereditary disease, including pedigree collection and optional genomic testing. Both \textbf{GPT-4.1} and \textbf{DeepSeek-V3} identify common issues such as informed consent, privacy protection, and genetic data security, but neither model raises concerns regarding \emph{misattributed parentage}---a well-documented risk in family-based genetic investigations. During the deliberation stage, Mirror’s \textbf{genetic-medicine expert} agent highlights the possibility that testing may reveal non-biological relationships (e.g., a child not genetically related to one stated parent). This issue does not appear explicitly in the dossier, and the PI is unable to provide a clear protocol for disclosure or psychological support if such findings occur. As a result, Mirror flags this as an unresolved ethical concern requiring clarification in subsequent revisions.
This case demonstrates that Mirror can surface implicit, context-sensitive ethical risks that are \emph{not} reducible to explicit rule violations, while other models focus mainly on generic consent and data issues. The multi-agent debate further validates the concern by revealing missing safeguards rather than asserting definitive non-compliance.

\paragraph{Case 2: Trial of a Food–Medicine Homology Formula.}
The second dossier evaluates a clinical efficacy study on a traditional “food–medicine homology’’ formula for treating a chronic condition. Both \textbf{GPT-4.1} and \textbf{DeepSeek-V3} conclude that the dossier presents no major ethical concerns, citing complete documentation, standard participant recruitment criteria, and appropriate safety monitoring procedures.
In contrast, Mirror’s \textbf{traditional-medicine expert} agent questions whether each herbal ingredient in the formula has undergone toxicity evaluation or risk classification for the target population. Although the formula is widely used and legally categorized as food–medicine homology, the dossier does not provide evidence addressing potential adverse interactions or contraindications. During debate, the PI is unable to supply ingredient-level safety justification, leading Mirror to mark this issue as unresolved and in need of supplementary pharmacology information.
This example illustrates Mirror’s ability to incorporate domain-specific expertise beyond standard biomedical protocols. The concern raised is not a regulatory violation but a request for additional scientific justification, aligning with real committee practices for early-stage traditional medicine trials.

Across both cases, Mirror exhibits two desirable properties. First, it uncovers latent ethical risks that are difficult to capture through rule-based methods, reflecting expert-level domain sensitivity. Second, through structured debate, Mirror distinguishes between (i) confirmed ethical problems and (ii) issues that stem from \emph{insufficient information}, thereby mirroring the behavior of real review committees. In contrast, GPT-4.1 and DeepSeek-V3 tend to provide high-level assessments and rarely challenge omissions in the dossier, leading to fewer but less comprehensive findings.
These case studies suggest that Mirror-Committee-Review can complement rule-based evaluation by providing deeper, context-aware ethical deliberation suitable for complex or emerging research domains.

\section{Limitations}

Although Mirror demonstrates strong potential for AI-assisted ethics review, several limitations remain. The training and evaluation corpora were primarily curated from publicly available materials, whereas authentic IRB archives, which contain richer contextual signals, deliberation dynamics, and institution-specific norms, are underrepresented. This reliance limits the system’s coverage of rare, highly contextual, or nuanced review scenarios.
The 100-case ethics-review benchmark is similarly skewed toward relatively standard life-science studies and does not yet capture the full breadth of ethics review settings, including high-risk clinical trials, multi-center studies, and sensitive social or behavioral research. In addition, EthicsLLM is built on an 8B-parameter backbone. While domain adaptation yields substantial performance gains, the model’s scale inherently constrains long-context reasoning and complex deliberative capabilities.
Future work will incorporate a broader range of IRB materials under stringent privacy and governance controls, expand the diversity and difficulty of benchmark cases, and explore larger or more advanced model architectures to further improve the rigor, robustness, and generalizability of AI-assisted ethics review.

\section{Conclusion}

In this work, we introduce \textbf{Mirror}, a multi-agent framework designed to support AI-assisted ethics review. Central to this framework is \textbf{EthicsLLM}, a domain-adapted language model fine-tuned on \textbf{EthicsQA}, a specialized dataset derived from authoritative normative and regulatory sources. This domain adaptation enables substantially more accurate, context-aware ethical reasoning than that of general-purpose large language models.
Mirror supports two complementary evaluation modes aligned with established institutional review practices: (1) \textbf{Mirror-ER}, which performs rule-based expedited review using an executable, expert-curated rule base; and (2) \textbf{Mirror-CR}, which extends the framework to full-board review by enabling committee-style deliberation through coordinated expert agents.
Empirical evaluations demonstrate that Mirror achieves strong performance across a range of tasks, including ethics QA, rule violation detection, and case-based committee assessment. Compared with general-purpose LLMs, Mirror produces judgments that are more consistent, interpretable, and contextually grounded, while also surfacing nuanced ethical considerations that go beyond explicit regulatory provisions. These results suggest that integrating domain-adapted models with structured rule representations and coordinated agentic workflows offers a practical and effective approach to AI-assisted ethics review.
This work opens several promising directions for future research, including the development of larger and more diverse benchmarks, extensions to additional regulatory and institutional contexts, and deeper investigation of human--AI collaborative decision-making. By releasing the datasets, model weights, and evaluation tools, we aim to support the broader development of reliable, transparent, and institution-ready AI systems for research governance.

\section{Ethics and Privacy Statement}

This research was conducted in accordance with applicable institutional and international ethical standards, with formal approval obtained from the relevant Ethics Committee. All data used in this project were collected with the informed consent of contributors.
\textbf{All training data used in the development of Mirror and EthicsLLM are synthetic} and contain no real personal, institutional, or proprietary information. These synthetic datasets were carefully designed to preserve the structural and statistical characteristics of authentic ethics-review materials, while maintaining strict separation from any original or sensitive sources. All data used for evaluation underwent a rigorous data de-identification process, consistent with established best practices and relevant regulatory guidelines.
For real-world deployment, we recommend that Mirror and its agentic components be operated within secure, institution-managed computing environments. Appropriate access controls, network isolation, and application-level safeguards should be implemented to prevent unauthorized access and mitigate the risk of privacy leakage.

\section*{Acknowledgments}
This work is in part supported by the National Natural Science Foundation of China (No. 62521004) and the Major Program of the National Social Science Foundation of China, “A Study of Bioethical Discourse from the Perspective of Practical Philosophy” (Grant No. 25\&ZD021).

\clearpage

\bibliographystyle{plainnat}
\bibliography{main}

@article{burman2001breaking,
  author  = {Burman, W J and Reves, R R and Cohn, D L and Schooley, R T},
  title   = {Breaking the camel's back: multicenter clinical trials and local institutional review boards},
  journal = {Ann Intern Med},
  year    = {2001},
  volume  = {134},
  number  = {2},
  pages   = {152--157},
  doi     = {10.7326/0003-4819-134-2-200101160-00016},
}

@article{cui2023chatlaw,
  author        = {Cui, J and Li, Z and Yan, Y and Chen, B and Yuan, L},
  title         = {ChatLaw: Open-Source Legal Large Language Model with Integrated External Knowledge Bases},
  journal       = {arXiv preprint arXiv:2306.16092},
  year          = {2023},
  eprint        = {2306.16092},
  archivePrefix = {arXiv},
  primaryClass  = {cs.CL},
  url           = {https://arxiv.org/abs/2306.16092},
  note          = {Version 1, 28 Jun 2023}
}

@article{yue2023disclawllm,
  author        = {Yue, S and Chen, W and Wang, S and Li, B and Shen, C and Liu, S and Zhou, Y and Xiao, Y and Yun, S and Huang, X and Wei, Z},
  title         = {DISC-LawLLM: Fine-tuning Large Language Models for Intelligent Legal Services},
  journal       = {arXiv preprint arXiv:2309.11325},
  year          = {2023},
  eprint        = {2309.11325},
  archivePrefix = {arXiv},
  primaryClass  = {cs.CL},
  url           = {https://arxiv.org/abs/2309.11325},
  note          = {Version 2, 23 Sep 2023}
}

@article{liu2021caseelements,
  author  = {Liu, H and Wang, L and Sun, Y and Chen, Y and Zhang, S and Lin, H},
  title   = {Case Element Recognition Method Based on Pre-trained Language Models},
  journal = {Journal of Chinese Information Processing},
  year    = {2021},
  volume  = {35},
  number  = {11},
  pages   = {91--100}
}

@article{zhang2021judgmentelements,
  author  = {Zhang, H and Pan, B and Zhang, Y},
  title   = {Judgment Element Extraction from Fact Descriptions in Legal Documents Based on Deep Learning},
  journal = {Computer Applications and Software},
  year    = {2021},
  volume  = {38},
  number  = {9},
  pages   = {160--166}
}

@inproceedings{janatian2023text2structure,
  author    = {Janatian, S and Westermann, H and Tan, J and Savelka, J and Benyekhlef, K},
  title     = {From Text to Structure: Using Large Language Models to Support the Development of Legal Expert Systems},
  booktitle = {Legal Knowledge and Information Systems - JURIX 2023: 36th Annual Conference},
  year      = {2023},
  editor    = {Sileno, G and Spanakis, J and van Dijck, G},
  series    = {Frontiers in Artificial Intelligence and Applications},
  volume    = {379},
  pages     = {167--176},
  publisher = {IOS Press},
  doi       = {10.3233/FAIA230962}
}

@article{goldman1982inconsistency,
  author  = {Goldman, J and Katz, M D},
  title   = {Inconsistency and Institutional Review Boards},
  journal = {JAMA},
  year    = {1982},
  volume  = {248},
  number  = {2},
  pages   = {197--202},
  doi     = {10.1001/jama.1982.03330020041027},
}

@article{mansbach2007variation,
  author  = {Mansbach, J and Acholonu, U and Clark, S and Camargo, C A, Jr.},
  title   = {Variation in Institutional Review Board Responses to a Standard, Observational, Pediatric Research Protocol},
  journal = {Acad Emerg Med},
  year    = {2007},
  volume  = {14},
  number  = {4},
  pages   = {377--380},
  doi     = {10.1197/j.aem.2006.11.031},
}

@article{green2006impact,
  author  = {Green, L A and Lowery, J C and Kowalski, C P and Wyszewianski, L},
  title   = {Impact of Institutional Review Board Practice Variation on Observational Health Services Research},
  journal = {Health Serv Res},
  year    = {2006},
  volume  = {41},
  number  = {1},
  pages   = {214--230},
  doi     = {10.1111/j.1475-6773.2005.00458.x},
}

@article{candilis2006need,
  author  = {Candilis, Philip J and Lidz, Charles W and Arnold, Robert M},
  title   = {The need to understand IRB deliberations},
  journal = {IRB},
  year    = {2006},
  volume  = {28},
  number  = {1},
  pages   = {1--5},
  pmid    = {16680872},
}

@misc{NHWC2023EthicsReviewEn,
  author = {{National Health Commission}; {Ministry of Education}; {Ministry of Science and Technology}; {National Administration of Traditional Chinese Medicine of the People’s Republic of China}},
  title  = {Measures for the Ethical Review of Life Science and Medical Research Involving Humans},
  year   = {2023},
  note   = {Joint regulatory measure (Guo Wei Ke Jiao Fa [2023] No.4)}
}

@article{porsdammann2025_irb_llm,
  author  = {Porsdam Mann, S and Seah, J J and Latham, S R and Savulescu, J and Aboy, M and Earp, B D},
  title   = {Chat-{IRB}? How application-specific language models can enhance research ethics review},
  journal = {J Med Ethics},
  year    = {2025},
  doi     = {10.1136/jme-2025-110845}
}

@article{sridharan2024_case,
  author  = {Sridharan, S and Sivaramakrishnan, S},
  title   = {Leveraging artificial intelligence to detect ethical concerns in medical research: A case study},
  journal = {J Med Ethics},
  year    = {2024},
  doi     = {10.1136/jme-2023-109235}
}

@article{sridharan2024_irb,
  author  = {Sridharan, S and Sivaramakrishnan, S},
  title   = {Assessing the decision-making capabilities of {AI} platforms as {IRB} members},
  journal = {J Empir Res Hum Res Ethics},
  year    = {2024},
  doi     = {10.1177/15562646241257288}
}

@inproceedings{governatori2006_compliance,
  author    = {Governatori, G and Milosevic, Z},
  title     = {A formal approach for legal compliance verification},
  booktitle = {Proceedings of the {IEEE} {EDOC} Conference},
  year      = {2006},
  pages     = {46--55},
  doi       = {10.1109/EDOC.2006.28}
}

@article{rojas2016_conformance,
  author  = {Rojas, E and Cabanillas, C and Mendling, J},
  title   = {Compliance checking in the construction industry: A process mining approach},
  journal = {Autom Constr},
  year    = {2016},
  volume  = {59},
  pages   = {1--15},
  doi     = {10.1016/j.autcon.2015.07.009}
}

@inproceedings{aires2018_embedding,
  author    = {Aires, J and de Freitas, R P O and Meneguzzi, F},
  title     = {A contextual embedding approach for automated contract clause classification},
  booktitle = {Proceedings of the International Conference on Autonomous Agents and Multiagent Systems (AAMAS)},
  year      = {2018},
  pages     = {1--9}
}

@article{huang2024_multitask,
  author  = {Huang, Y and Liu, Q and Li, M},
  title   = {Multi-task learning for legal clause classification and compliance assessment},
  journal = {Artif Intell Law},
  year    = {2024},
  note    = {Online first}
}

@article{lippi2019_claudette,
  author  = {Lippi, M and Torroni, P},
  title   = {Claudette: An automated system for detecting potentially unfair clauses in online terms of service},
  journal = {Artif Intell Law},
  year    = {2019},
  volume  = {27},
  pages   = {117--139},
  doi     = {10.1007/s10506-018-9231-z}
}

@inproceedings{sun2025_rag_compliance,
  author    = {Sun, L and Yang, Y and Joty, S},
  title     = {A compliance checking framework based on retrieval-augmented generation},
  booktitle = {Proceedings of the International Conference on Computational Linguistics (COLING)},
  year      = {2025}
}

@inproceedings{gema2025_mmlu,
  author    = {Gema, A P and Leang, J O J and Hong, G and Devoto, A and Mancino, A C M and Saxena, R and He, X and Zhao, Y and Du, X and Ghasemi Madani, M R and Barale, C and McHardy, R and Harris, J and Kaddour, J and Van Krieken, E and Minervini, P},
  title     = {Are We Done with MMLU?},
  booktitle = {Proceedings of the 2025 Conference of the Nations of the Americas Chapter of the Association for Computational Linguistics: Human Language Technologies (NAACL-HLT)},
  year      = {2025}
}

@article{singhal2024_medpalm2,
  author  = {Singhal, K and Tu, T and Gottweis, J and et al.},
  title   = {Toward expert-level medical question answering with large language models},
  journal = {Nat Med},
  year    = {2024},
  volume  = {31},
  pages   = {943--950},
  doi     = {10.1038/s41591-024-03423-7}
}

@article{zhang2023_huatuogpt2,
  author  = {Zhang, A and et al.},
  title   = {HuatuoGPT-{II}: One-stage training for medical adaptation of large language models in Chinese medicine},
  journal = {arXiv preprint},
  year    = {2023},
  note    = {arXiv:2311.09774}
}

@article{cui2023_chatlaw,
  author  = {Cui, J and Li, Z and Yan, Y and Chen, B and Yuan, L},
  title   = {ChatLaw: Open-source legal large language model with integrated external knowledge bases},
  journal = {arXiv preprint},
  year    = {2023},
  note    = {arXiv:2306.16092}
}

@article{yue2023_disc_lawllm,
  author  = {Yue, S and Chen, W and Wang, S and et al.},
  title   = {{DISC}-LawLLM: Fine-tuning large language models for intelligent legal services},
  journal = {arXiv preprint},
  year    = {2023},
  note    = {arXiv:2309.11325}
}

@inproceedings{li2023camel,
  author    = {Li, G and Hammoud, H A K and Itani, H and Khizbullin, D and Ghanem, B},
  title     = {{CAMEL}: Communicative agents for ``mind'' exploration of large language model society},
  booktitle = {Advances in Neural Information Processing Systems},
  year      = {2023},
  volume    = {36}
}

@inproceedings{du2024multiagent,
  author    = {Du, Y and Li, S and Torralba, A and Tenenbaum, J B and Mordatch, I},
  title     = {Improving factuality and reasoning in language models through multiagent debate},
  booktitle = {Proceedings of the International Conference on Machine Learning},
  year      = {2024},
  volume    = {235},
  pages     = {11733--11763}
}

@article{ma2025multiagent,
  author  = {Ma, Z and Bahja, A R and Burgdorf, A and Pomp, A and Meisen, T and Jorgensen, B N and Ma, Z G},
  title   = {Multi-agent multimodal large language model framework for automated interpretation of fuel efficiency analytics in public transportation},
  journal = {Appl Sci},
  year    = {2025},
  volume  = {15},
  number  = {21},
  pages   = {11619},
  doi     = {10.3390/app152111619}
}

@article{huang2023ceval,
  author  = {Huang, Y and Bai, Y and Zhu, Z and Zhang, J and Zhang, J and Su, T and Liu, J and Lv, C and Zhang, Y and Lei, J and Fu, Y and Sun, M and He, J},
  title   = {{C}-{E}val: A multi-level multi-discipline Chinese evaluation suite for foundation models},
  journal = {arXiv preprint},
  year    = {2023},
  note    = {arXiv:2305.08322}
}

% \clearpage

% \newpage

% \beginappendix

% \startcontents[app]
% \begingroup
%   \renewcommand{\contentsname}{Appendix Contents}
%   \section*{\contentsname}
%   \printcontents[app]{}{1}{}
% \endgroup
% \newpage

% \input{section/appendix}

\end{document}